\documentclass[10pt,twocolumn,letterpaper]{article}

\usepackage{iccv}
\usepackage{times}
\usepackage{epsfig}
\usepackage{graphicx}
\usepackage{amsmath}
\usepackage{amssymb}
\usepackage{algorithm}
\usepackage{algpseudocode}
\usepackage{caption}
\usepackage{multirow}

\usepackage[accsupp]{axessibility}

\usepackage[pagebackref=true,breaklinks=true,colorlinks,bookmarks=false]{hyperref}

\iccvfinalcopy 

\def\iccvPaperID{2388} 

\ificcvfinal\fi

\begin{document}

\title{Delicate Textured Mesh Recovery from NeRF via Adaptive Surface Refinement}

\author{
Jiaxiang Tang$^1$, 
Hang Zhou$^2$, 
Xiaokang Chen$^1$, 
Tianshu Hu$^2$, 
Errui Ding$^2$, 
Jingdong Wang$^2$,
Gang Zeng$^1$\\
$^1$Key Lab. of Machine Perception (MoE), School of IST, Peking University. \quad $^2$Baidu Inc.\\
{\tt\small \{tjx, pkucxk, zeng\}@pku.edu.cn} \quad {\tt\small \{zhouhang09,hutianshu01,dingerrui,wangjingdong\}@baidu.com} \\
}

\twocolumn[{
\renewcommand\twocolumn[1][]{#1}
\maketitle
\vspace{-1cm}
\centering
\captionsetup{type=figure}
\includegraphics[width=\textwidth]{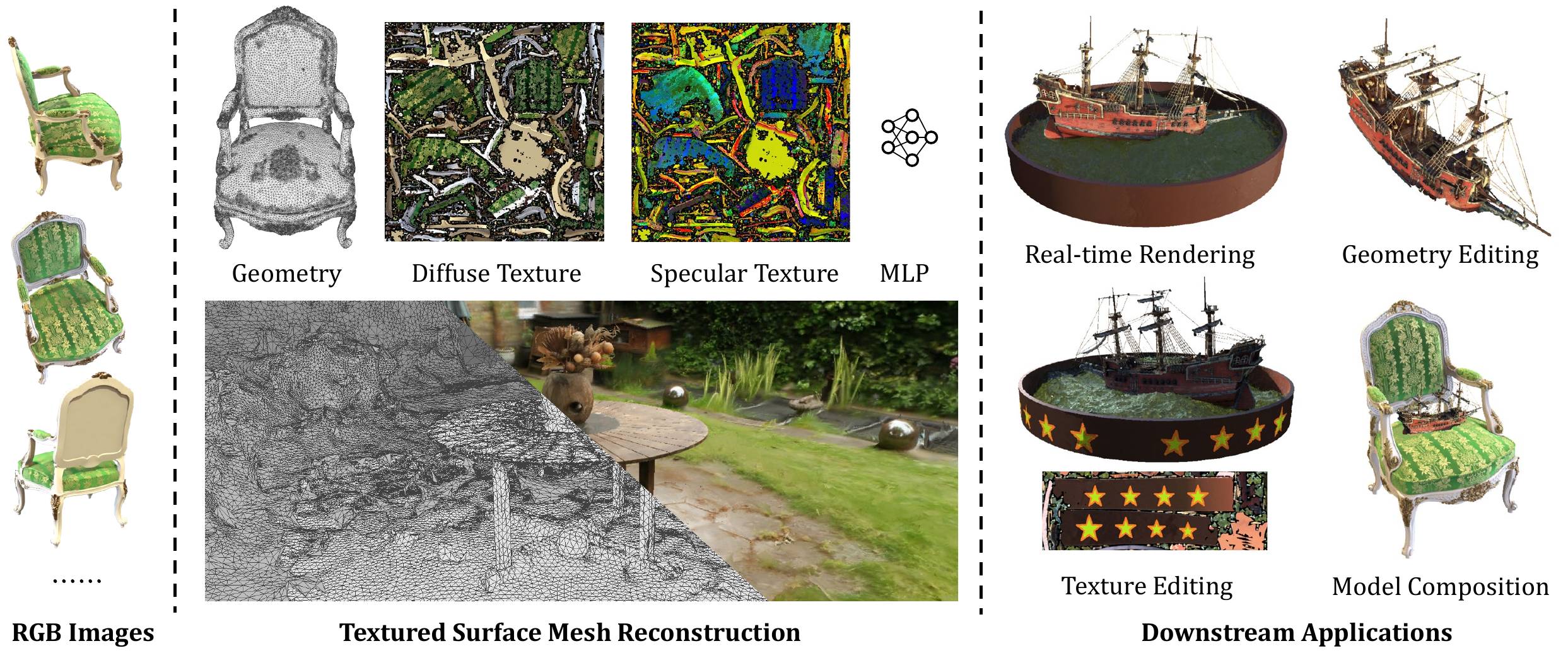}
\captionof{figure}{
Our framework, \textit{NeRF2Mesh}, reconstructs high-quality surface meshes with diffuse and specular textures from multi-view RGB images, generalizing well from object- to scene-level datasets. 
The exported textured meshes are ready-to-use for common graphics hardware and software, facilitating various downstream applications.
}
\label{fig:teaser}
\vspace{0.5cm}
}]

\maketitle
\ificcvfinal\thispagestyle{empty}\fi

\begin{abstract}
\vspace{-0.2cm}
Neural Radiance Fields (NeRF) have constituted a remarkable breakthrough in image-based 3D reconstruction. 
However, their implicit volumetric representations differ significantly from the widely-adopted polygonal meshes and lack support from common 3D software and hardware, making their rendering and manipulation inefficient.
To overcome this limitation, we present a novel framework that generates textured surface meshes from images. 
Our approach begins by efficiently initializing the geometry and view-dependency decomposed appearance with a NeRF.
Subsequently, a coarse mesh is extracted, and an iterative surface refinement algorithm is developed to adaptively adjust both vertex positions and face density based on re-projected rendering errors. 
We jointly refine the appearance with geometry and bake it into texture images for real-time rendering.
Extensive experiments demonstrate that our method achieves superior mesh quality and competitive rendering quality.
\end{abstract}

\section{Introduction}
The reconstruction of 3D scenes from RGB images is a complex task in computer vision with many real-world applications. 
In recent years, Neural Radiance Fields (NeRF)~\cite{mildenhall2020nerf,barron2021mip,TensoRF,mueller2022instant} have gained popularity
for their impressive ability to reconstruct and render large-scale scenes with realistic details.
However, NeRF representations often use implicit functions and specialized ray marching algorithms for rendering, making them difficult to manipulate and slow to render due to poor hardware support, which limits their use in downstream applications.
In contrast, polygonal meshes are the most commonly used representation in 3D applications and are well-supported by most graphic hardware to accelerate rendering. 
However, direct reconstruction of meshes can be challenging due to their irregularity, and most approaches are limited to
object-level reconstructions~\cite{munkberg2022extracting,chen2019learning,chen2021dib}.

Some recent works~\cite{munkberg2022extracting,boss2022samurai,chen2022mobilenerf,bakedsdf} have focused on combining the advantages of both NeRF and mesh representation.
MobileNeRF~\cite{chen2022mobilenerf} presents a method of optimizing NeRF on a grid mesh and incorporates rasterization for real-time rendering. 
However, the resulting mesh is far from the real surface of the reconstructed scene.
Besides, the textures are in the feature space instead of the RGB space, which makes editing or manipulation inconvenient.
To obtain accurate surface meshes, a popular approach is to use Signed Distance Fields (SDF), which defines an exact surface~\cite{wang2021neus,yariv2021volume,yu2022monosdf}.
However, this line of research typically generates over-smoothed geometry that fails to model thin structures.
Additionally, meshes obtained through Marching Cubes~\cite{lorensen1987marching} produce a large number of redundant vertices and faces to keep details.
NVdiffrec~\cite{munkberg2022extracting} uses a differentiable rasterizer~\cite{Laine2020diffrast} to optimize a deformable tetrahedral grid but is limited to object-level reconstruction and also fails to recover complex topology.
The presence of a representation gap makes it challenging to recover accurate surface meshes from volumetric NeRF while maintaining rendering quality.

This paper presents a novel framework called \textit{NeRF2Mesh} for extracting delicate textured surface meshes from RGB images, as illustrated in Figure~\ref{fig:teaser}. 
Our key insight is to \textbf{refine a coarse mesh extracted from NeRF for joint optimization of geometry and appearance}.
The volumetric NeRF representation is suitable for efficient initialization of geometry and appearance.
With a coarse mesh extracted from NeRF, we adjust the vertices' position and face density based on 2D rendering errors, which in turn contributes to appearance optimization.
To enable texture editing, we decompose the appearance into view-independent diffuse and view-dependent specular terms, so the diffuse color can be exported as a standard RGB image texture. 
The specular term is exported as a feature texture that produces view-dependent color through a small MLP embedded in the fragment shader.
Overall, our framework enables the creation of versatile and practical mesh assets that can be used in a range of scenarios that are challenging for volumetric NeRF.

Our contributions can be summarized as follows: 
\begin{itemize}
\item We present the {NeRF2Mesh} framework to reconstruct textured surface meshes from multi-view RGB images, by jointly refining the geometry and appearance of coarse meshes extracted from an appearance decomposed NeRF. 
\item We propose an iterative mesh refinement algorithm that enables us to adaptively adjust face density, where complex surfaces are subdivided and simpler surfaces are decimated based on re-projected 2D image errors.
\item Our method achieves enhanced surface mesh quality, relatively smaller mesh size, and competitive rendering quality to recent methods. Furthermore, the resulting meshes can be real-time rendered and interactively edited with common 3D hardware and software.
\end{itemize}

\section{Related Work}
\subsection{NeRF for Scene Reconstruction}
NeRF~\cite{mildenhall2020nerf} and its subsequent works~\cite{barron2021mip,barron2022mipnerf360,nerf++,kilonerf,xiang2021neutex,park2021nerfies,xiang2021neutex,attal2023hyperreel,cao2022real,reiser2023merf} represent a remarkable advancement in 3D scene reconstruction from RGB images. 
Despite the superior rendering quality, vanilla NeRF faces several issues. 
For instance, the model's training and inference speed is slow due to the large number of MLP evaluations, which limits the widespread adoption of NeRF representation. 
To address this, several works~\cite{yu2021plenoctrees,yu_and_fridovichkeil2021plenoxels,sun2021direct,mueller2022instant,TensoRF} have proposed methods to reduce the MLP's size or eliminate it altogether, and instead optimize an explicit 3D feature grid that stores the density and appearance information. 
DVGO~\cite{sun2021direct} employs two dense feature grids for density and appearance encoding, but the dense grid leads to a large model size. 
To effectively control the model size, Instant-NGP~\cite{mueller2022instant} proposes a multi-resolution hash table.
In addition to the efficiency issue, the implicit representation of NeRF cannot be directly manipulated and edited in both geometry and appearance, unlike explicit representations such as polygonal meshes. 
Although some works~\cite{lazova2022control,tang2022compressible,yang2021objectnerf,liu2021editing,wang2021clip} explore geometry manipulation and composition of NeRF, they are still limited in different ways. 
On the other hand, others~\cite{zhang2021nerfactor,boss2021nerd,physg2021,srinivasan2021nerv,bi2020neural,verbin2021refnerf} aim to decompose the reflectance under unknown illumination to enable relighting and texture editing. 
These problems result in a gap between NeRF representation and widely used polygonal meshes in downstream applications. 
Our objective is to narrow this gap by exploring methods to convert NeRF reconstructions into textured meshes.

\subsection{Surface Mesh for Scene Reconstruction}

Reconstructing explicit surface meshes directly can be challenging, particularly for complex scenes with intricate topology. 
Most approaches in this area of research assume a template mesh with a fixed topology~\cite{chen2019learning,chen2021dib,jatavallabhula2019kaolin,liu2019soft}. 
Recent methods~\cite{munkberg2022extracting,hasselgren2022nvdiffrecmc,liao2018deep,shen2021deep} have begun to address topology optimization.
NVdiffrec~\cite{munkberg2022extracting} combines differentiable marching tetrahedrons~\cite{liao2018deep} with differentiable rendering to optimize surface meshes directly. 
It can also decompose materials and illumination, which is further improved in NVdiffrecMC~\cite{hasselgren2022nvdiffrecmc} using Monte Carlo rendering.
Nonetheless, these methods still have limitations in that they only apply to object-level mesh reconstruction and struggle to differentiate between background and foreground meshes in unbounded outdoor scenes. 
A foreground mask~\cite{munkberg2022extracting} must be prepared to optimize the object boundary using differentiable rendering. 
In contrast, our focus is on surface mesh reconstruction at both the object and scene levels without the need of semantic masks.

\subsection{Extracting Surface Mesh from NeRF}
NeRF represents geometry using a volumetric density field, which may not necessarily form a concrete surface. 
To address this, a popular strategy is to learn a SDF~\cite{wang2021neus,yariv2021volume,darmon2022improving,fu2022geo,yu2022monosdf,wu2022voxurf,yang2022neumesh}, where the surface can be determined by the zero level set. 
NeuS~\cite{wang2021neus} applies a SDF to density transformation to enable differentiable rendering, and the Marching Cubes~\cite{lorensen1987marching} algorithm is usually used to extract the surface mesh from these volumes. 
BakedSDF~\cite{bakedsdf} optimize a hybrid SDF volume-surface representation and bake it into meshes for real-time rendering.
However, SDF-based methods tend to learn over-smoothed geometry and fail to handle thin structures.
Some methods~\cite{ueda2022neural,long2022neuraludf} explore Unsigned Distance Field (UDF) or a combination of density field and SDF to address this limitation, but they are still limited to object-level reconstruction.
SAMURAI~\cite{boss2022samurai} aims to jointly recover camera poses, geometry, and appearance of a single object under unknown captured conditions and export textured meshes. 
MobileNeRF~\cite{chen2022mobilenerf} proposes to train NeRF on a grid mesh, which can be rendered in real-time. 
However, their mesh is not exactly the surface mesh and only exports features as texture, which have to be rendered with a custom shader and are unfriendly for editing.
Recent works~\cite{mueller2022instant,poole2022dreamfusion} have found that an exponential density activation can help to concentrate the density and form better surfaces. 
We also adopt density field to capture correct topology, since the surface can be further refined.

\section{Method}
\begin{figure*}[t!]
    \centering
    \includegraphics[width=\textwidth]{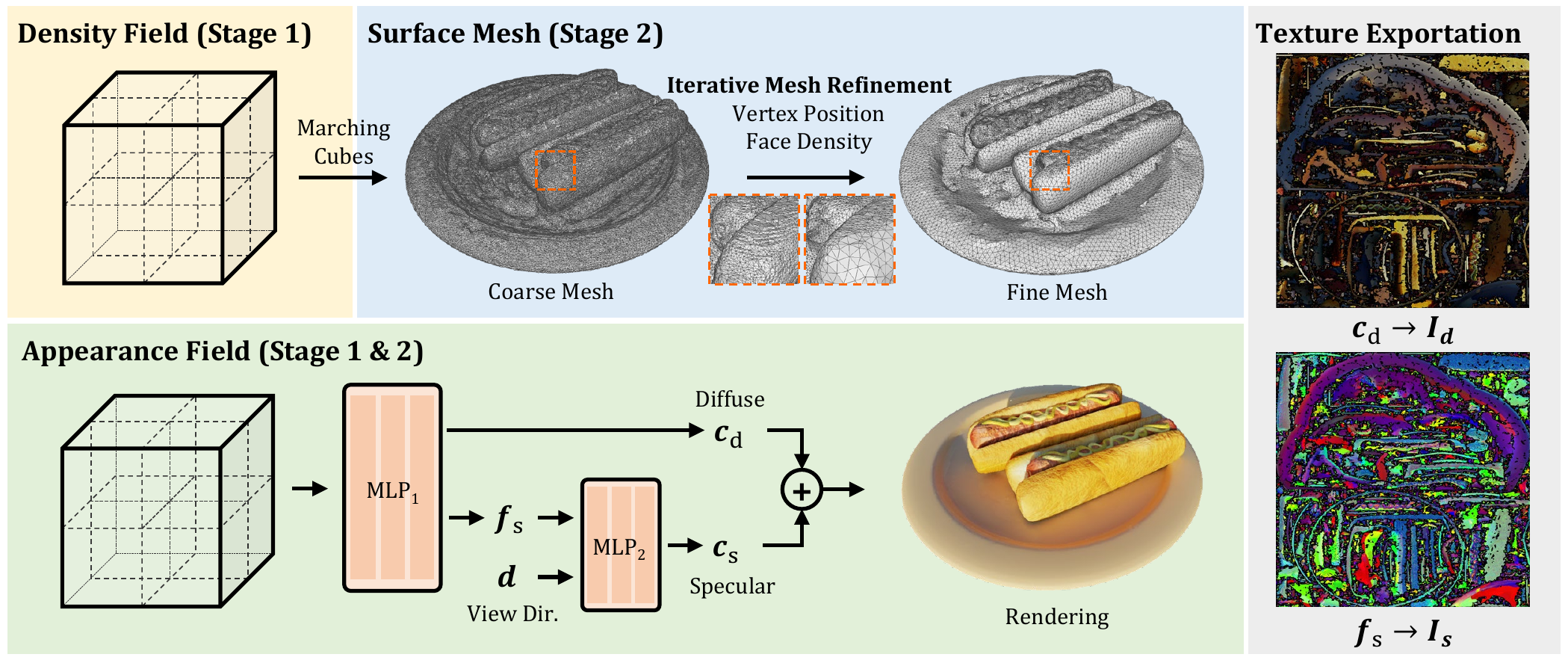}
    \caption{
    \textbf{NeRF2Mesh Framework}.
    The geometry is initially learned with a density grid, and then extracted to form a coarse mesh. 
    We optimize it into a fine mesh with more accurate surface and adaptive face density.
    The appearance is learned with a color grid shared by two stages, and decomposed into diffuse and specular terms.
    After convergence, we can export the fine mesh, unwrap its UV coordinates, and bake the appearance into texture images.
    }
    \label{fig:network}
\end{figure*}

In this section, we introduce our framework, as shown in Figure~\ref{fig:network}, for reconstructing a textured surface mesh from a collection of RGB images that is compatible with common 3D hardware and software. 
The training process comprises two stages. 
Firstly, we train a grid-based NeRF~\cite{mueller2022instant} to efficiently initialize the geometry and appearance of the mesh (Section~\ref{sec:stage1}). 
Next, we extract a coarse surface mesh and fine-tune both surface geometry and appearance (Section~\ref{sec:stage2}). 
Once the training is complete, we can export a textured surface mesh in standard formats such as wavefront OBJ and PNG, which is ready-to-use for various downstream applications (Section~\ref{sec:export}).

\subsection{Efficient NeRF Training (Stage 1)}
\label{sec:stage1}

In the first stage, we leverage the volumetric NeRF representation to recover both the geometry and appearance of arbitrary scenes. 
The primary goal of this stage is to \textit{efficiently establish topologically accurate geometry and decomposed appearance in preparation for the subsequent surface mesh refinement phase}. 
While direct work on polygonal meshes~\cite{munkberg2022extracting} presents challenges in learning complex geometries, volumetric NeRF~\cite{mildenhall2020nerf} provides a more accessible alternative. 
We follow recent advancements in grid-based NeRF~\cite{mueller2022instant, sun2021direct, TensoRF, yu_and_fridovichkeil2021plenoxels} to enhance the efficiency of NeRF by employing two separate feature grids to represent the 3D space.

\noindent \textbf{Geometry.}
Geometry learning is facilitated through a density grid~\cite{mueller2022instant} and a shallow MLP expressed as follows:
\begin{equation}
    \sigma = \phi(\text{MLP}(E^\text{geo}(\mathbf{x}))),
\end{equation}
where $\phi$ is the exponential activation~\cite{mueller2022instant} that promotes sharper surface, $E^\text{geo}$ is a learnable multi-resolutional feature grid, and $\mathbf{x} \in \mathbb R^3$ is the position of any 3D point.

\begin{figure}[t]
    \centering
    \includegraphics[width=\linewidth]{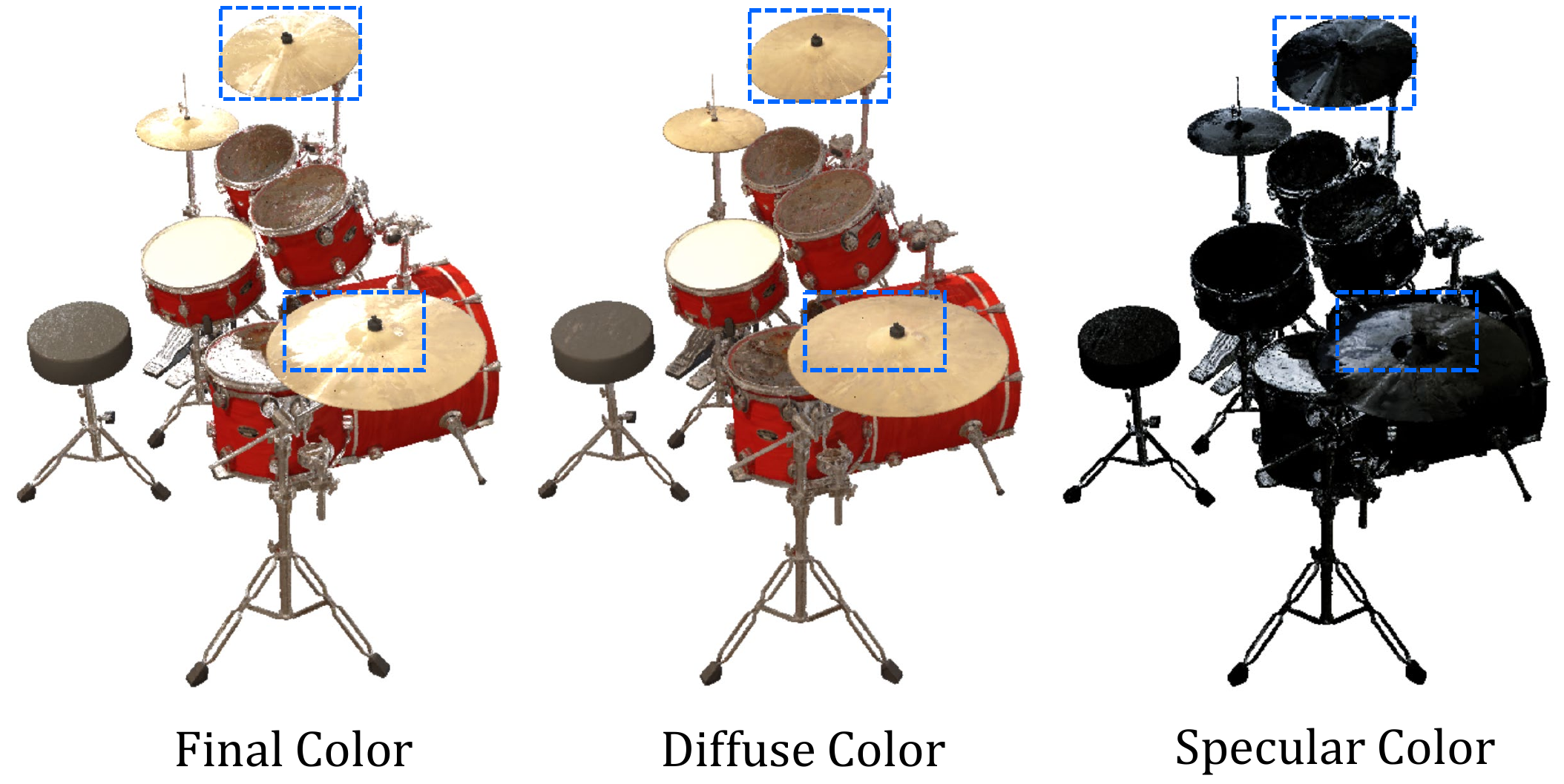}
    \caption{\textbf{Appearance Decomposition}.
    We separately model the diffuse and specular color.
    }
    \vspace{-0.3cm}
    \label{fig:diffuse_separation}
\end{figure}

\noindent \textbf{Appearance Decomposition.}
NeRF typically operates under no assumption of illumination or material properties.
As such, previous works have mainly employed a 5D implicit function conditioned on 3D position and 2D view direction to model view-dependent appearance. 
This approach renders the appearance as a black box, making it challenging to represent appearance with traditional 2D texture images.

To address this issue, we decompose the appearance into view-independent diffuse color $\mathbf{c}_d$ and view-dependent specular color $\mathbf{c}_s$ using a color grid and two shallow MLPs, expressed as follows:
\begin{gather}
    \mathbf{c}_d, \mathbf{f}_s = \psi(\text{MLP}_{1}(E^\text{app}(\mathbf{x}))), \\
    \mathbf{c}_s = \psi(\text{MLP}_{2}(\mathbf{f}_s, \mathbf{d})),
\end{gather}
where $\psi$ refers to the sigmoid activation, $\mathbf{f}_s$ represents the intermediate features for the specular color at position $\mathbf{x}$, and $\mathbf{d}$ represents the view direction. 
The final color is obtained by summing the two terms:
\begin{equation}
    \mathbf{c} = \mathbf{c}_d + \mathbf{c}_s,
\end{equation}

As shown in Figure~\ref{fig:diffuse_separation}, we successfully separate the diffuse and specular terms. 
The diffuse color in $\mathbb R^3$ can be directly converted to an RGB image texture. 
Meanwhile, the specular features $\mathbf{f}_s$ can also be converted to textures, and the small $\text{MLP}_2$ can be fit into a fragment shader following~\cite{chen2022mobilenerf}.
Consequently, the specular color can also be exported and rendered later (see Section~\ref{sec:export} for details).

Our approach involves baking the lighting conditions into the textures. 
This is because estimating the environment lighting can be challenging for realistic datasets, and previous studies have observed that this can result in reduced rendering quality~\cite{zhang2021nerfactor,munkberg2022extracting}.

\noindent \textbf{Loss function.}
To optimize our model, we use the original NeRF's rendering loss. 
Given a ray $\mathbf{r}$ originating from $\mathbf{o}$ with direction $\mathbf{d}$, we query the model at positions $\mathbf{x}_i = \mathbf{o} + t_i \mathbf{d}$, sequentially sampled along the ray, for densities ${\sigma_i}$ and colors ${\mathbf{c}_i}$. 
The final pixel color is obtained by numerical quadrature using the following equation:
\begin{gather}
\label{eq:volume_rendering}
    \mathbf{\hat C}(\mathbf{r}) = \sum_i T_i \alpha_i \mathbf{c}_i, 
    T_i = \prod_{j < i} (1 - \alpha_j),
\end{gather}
where $\delta_i = t_{i+1} - t_i$ is the step size, $\alpha_i = 1 - \exp(-\sigma_i \delta_i)$ is the point-wise rendering weight, and $T_i$ is the transmittance. 
We minimize the loss between each pixel's predicted color $\mathbf{\hat C}(\mathbf{r})$ and the ground truth color $\mathbf{C}(\mathbf{r})$:
\begin{equation}
    \label{loss:main}
    \mathcal{L}_{\text{NeRF}} = \sum_{\mathbf{r}} || \mathbf{C}(\mathbf{r}) - \mathbf{\hat C}(\mathbf{r}) ||^2,
\end{equation}
We encourage separation of the diffuse and specular terms by applying a L2 regularization on the specular color:
\begin{equation}
    \label{loss:sepc}
    \mathcal{L}_{\text{specular}} = \sum_{i} || \mathbf{c}_s(\mathbf{x}_i) ||^2,
\end{equation}
To make the surface sharper, we apply an entropy regularization on the rendering weights:
\begin{equation}
    \mathcal{L}_\text{entropy} = - \sum_{i} (\alpha_i \log \alpha_i + (1 - \alpha_i) \log (1 - \alpha_i) )
\end{equation}
where $\alpha_i$ is the per-point rendering weight. 
For unbounded outdoor scenes, we also apply Total Variation (TV) regularization on the density field $E^\text{geo}$ to reduce floaters~\cite{sun2021direct,TensoRF}.


\begin{figure}[t]
    \centering
    \includegraphics[width=\linewidth]{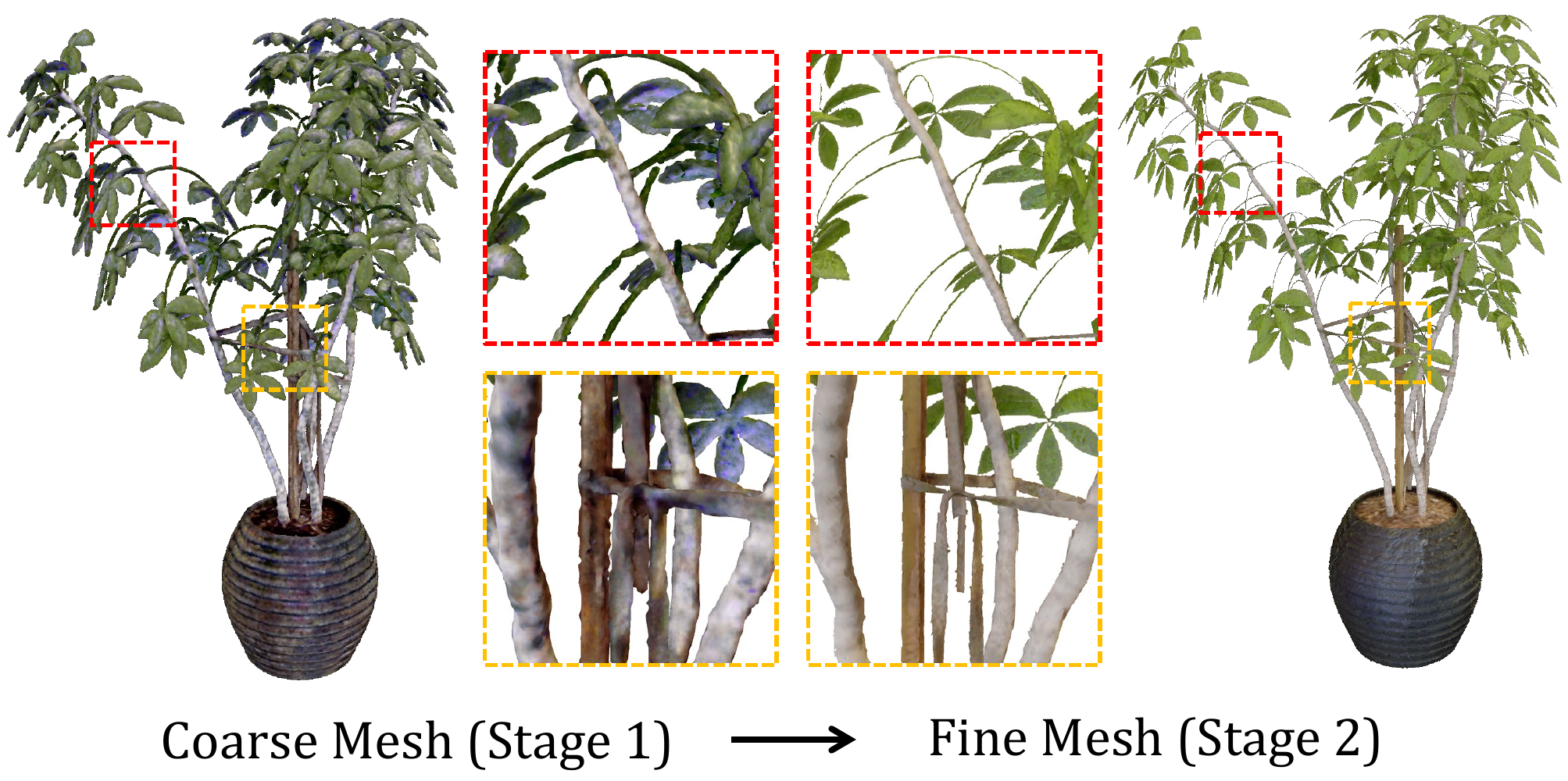}
    \caption{
    \textbf{Mesh Refinement}.
    We refine both the geometry and appearance of the coarse mesh in stage 2.
    }
    \vspace{-0.3cm}
    \label{fig:refine}
\end{figure}

\subsection{Surface Mesh Refinement (Stage 2)}
\label{sec:stage2}

After stage 1 converges, we apply Marching Cubes~\cite{lorensen1987marching} to extract a coarse mesh $\mathcal M_\text{coarse}$ from the density field, which serves as the initialization for stage 2.
We combine differentiable rendering technique with an iterative surface refinement algorithm. 
The goal of this stage is to accomplish \textit{joint optimization of the geometry and appearance for the coarse mesh extracted in the first stage}.
The process involves enhancement of the vertex positions, face density, and surface appearance, as depicted in Figure~\ref{fig:refine}.

\noindent \textbf{Appearance refinement.}
We use \texttt{nvdiffrast}~\cite{munkberg2022extracting} to perform differentiable rendering.
The mesh undergoes rasterization and the 3D positions are interpolated onto the image space pixel-wisely. 
Since the pixel colors are still queried in a point-wise manner, the appearance model from stage 1 can be inherited into stage 2. 
This eliminates the need to learn the appearance from scratch, reducing the required training steps for stage 2 to converge.
The pixel-wise color loss in Equation~\ref{loss:main} is still applied in stage 2 to allow joint optimization of appearance and geometry.

\noindent \textbf{Iterative mesh refinement.}
The coarse mesh $\mathcal M_\text{coarse}$ are often flawed.
These flaws include inaccurate vertices and dense, evenly-distributed faces, leading to vast disk storage and slow rendering speed.
Our goal is to recover delicate meshes resembling human-made ones by refining both vertex positions and face density.

Given an initial coarse mesh $\mathcal M_\text{coarse} = \{\mathcal V, \mathcal F\}$, we assign a trainable offset $\mathbf{\Delta v}_i$ to each vertex $\mathbf{v}_i \in \mathcal V$.
We use differentiable rendering~\cite{Laine2020diffrast} to optimize these offsets by back-propagating the image-space loss gradients~\cite{munkberg2022extracting}.
In contrast, mesh faces are not differentiable and cannot be optimized via back-propagation in the same way. 
To address this problem, we propose an iterative mesh refinement algorithm, which is inspired by the Iteratively Reweighted Least Squares (IRLS) algorithm~\cite{holland1977robust}. 
The key idea is to adaptively adjust face density based on previous training errors, given that inaccurate surface is among the factors contributing to large rendering error.
During training, we re-project the 2D pixel-wise rendering errors from Equation~\ref{loss:main} to the corresponding mesh faces and accumulate face-wise errors. 
After a certain number of iterations, we sort all face errors $E_\text{face}$ and determine two thresholds: 
\begin{gather}
    e_\text{subdivide} = \text{percentile}(E_\text{face}, 95), \\
    e_\text{decimate} = \text{percentile}(E_\text{face}, 50),
\end{gather}
Faces with error above $e_\text{subdivide}$ are mid-point subdivided~\cite{meshlab} to increase face density, while faces with error below $e_\text{decimate}$ are decimated~\cite{garland1997surface} and remeshed to reduce face density.
After the mesh updating, we reinitialize the vertex offsets and face errors and continue the training. 
This process is repeated several times until stage 2 finishes.

\noindent \textbf{Unbounded scene.}
Without loss of generality, we are able to model forward-facing~\cite{mildenhall2019llff} and unbounded outdoor scenes~\cite{barron2022mipnerf360}.
We divide the scene into multiple geometrically growing regions $[-2^k, 2^k]^3, k \in \{0, 1, 2, \cdots\}$ similar to Instant-NGP~\cite{mueller2022instant}.
Each region exports a separate mesh, with the overlapping part automatically excluded to form the complete geometry of the scene.
As the outer regions ($k \ge 1$) have a lower level of detail in comparison to the centre region ($k = 0$), we reduce the marching cubes resolution as $k$ increases. 
The iterative mesh refinement procedure solely focuses on the center region since the outer regions have relatively simpler geometry.

\noindent \textbf{Loss function.}
To prevent abrupt geometry, we apply a Laplacian smoothing loss $\mathcal L_\text{smooth}$~\cite{nealen2006laplacian,ravi2020pytorch3d}:
\begin{equation}
    \label{loss:smooth}
    \mathcal L_\text{smooth} = \sum_i \sum_{j \in S_i} \frac {1} {|S_i|} ||(\mathbf{v}_i + \mathbf{\Delta v}_i) - (\mathbf{v}_j + \mathbf{\Delta v}_j)||^2,
\end{equation}
where $S_i$ is the set of neighboring vertex indices for $\mathbf{v}_i$.
In addition, we regularize the vertices offset with an L2 loss:
\begin{equation}
    \label{loss:offset}
    \mathcal L_\text{offset} = \sum_i ||\mathbf{\Delta v}_i||^2,
\end{equation}
This ensures that the vertices do not move too far from their original positions.

\subsection{Mesh Exportation}
\label{sec:export}

The ultimate goal of our framework is to \textit{export a surface mesh with textures that are compatible with commonly used 3D hardware and software}. 
We currently have a surface mesh $\mathcal M_\text{fine}$ from stage 2, but the appearance is still encoded in a 3D color grid. 
To extract the appearance as texture images, we first unwrap the UV coordinates of $\mathcal M_\text{fine}$~\cite{xatlas}. 
Subsequently, we bake the surface's diffuse color $\mathbf{c}_d$ and specular features $\mathbf{f}_s$ into two separate images, $I_d$ and $I_s$, respectively.

\noindent \textbf{Real-time rendering.}
Our exported mesh can be efficiently accelerated and rendered in real-time, as a conventional textured mesh. 
The diffuse texture $I_d$ can be interpreted as an RGB image and rendered in most OpenGL-enabled devices with 3D software packages (\textit{e.g.}, Blender~\cite{blender}, Unity~\cite{unity}). 
To render the specular color, we adopt the approach proposed in MobileNeRF~\cite{chen2022mobilenerf}. 
We export the weights of the small $\text{MLP}_2$ and incorporate them into a fragment shader. 
This custom shader enables real-time evaluation and the addition of the specular term to the diffuse term, allowing for view-dependent effects.

\noindent \textbf{Mesh manipulation.}
Similar to a conventional textured mesh, the mesh we export can be readily modified and edited in terms of both geometry and appearance. 
Additionally, it facilitates the combination of multiple exported meshes, as can be observed in Figure~\ref{fig:teaser}.

\section{Experiment}
\begin{figure*}[t]
    \centering
    \includegraphics[width=\linewidth]{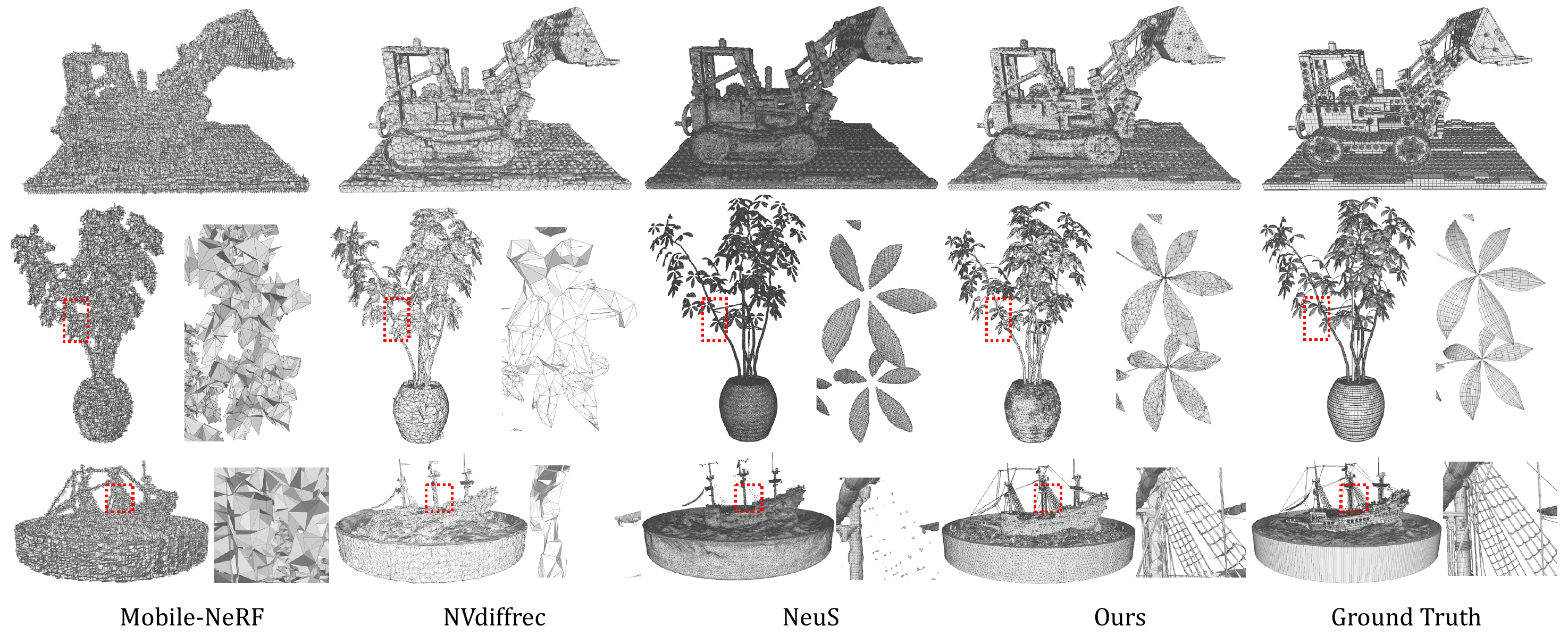}
    \caption{
    \textbf{Surface reconstruction quality on NeRF-synthetic dataset}.
    Our method achieves superior mesh reconstruction quality compared to previous methods, especially on thin structures with complex topology.
    We decimate meshes from NeuS~\cite{wang2021neus} to 25\% of the original faces since they are too dense to visualize.
    }
    \label{fig:mesh_quality}
\end{figure*}

\begin{table*}[!t]
\begin{center}

\begin{tabular}{l|cccccccc|c}
\hline
                                       & Chair & Drums & Ficus & Hotdog & Lego & Materials & Mic  & Ship & Mean \\
\hline
NeuS~\cite{wang2021neus}               & \textbf{3.95}  & 6.68  & 2.84  & 8.36   & 6.62 & \textbf{4.10}      & \textbf{2.99} &  9.54 & 5.64 \\
NVdiffrec~\cite{munkberg2022extracting}& 4.13           & 8.27  & 5.47  & 7.31   & \textbf{5.78} & 4.98      & 3.38 & 25.89 & 8.15 \\
\hline
Ours (coarse mesh)                     & 5.76           & 7.81  & 6.05  & 7.09   & 7.15 & 4.95      & 8.71 & 10.32 & 7.23 \\
Ours (fine mesh)                       & 4.60           & \textbf{6.02}  & \textbf{2.44}  & \textbf{5.19}   & 5.85 & 4.51      & 3.47 &  \textbf{8.39} & \textbf{5.06} \\

\hline
\end{tabular}

\end{center}
\caption{
\textbf{Chamfer Distance $\downarrow$ (Unit is $10^{-3}$)} on the NeRF-synthetic dataset compared to the ground truth meshes.
}
\label{tab:cd}
\end{table*}

\subsection{Implementation Details}

In the first stage, we train for $30,000$ steps, with each step evaluating approximately $2^{18}$ points. 
An exponentially decayed learning rate schedule ranging from $1\times 10^{-2}$ to $1\times 10^{-3}$ is employed. 
Specifically, during the initial $1,000$ steps, training solely employs the diffuse color to encourage the appearance factorization.
For the second stage, we train additional $10,000$ to $30,000$ steps based on convergence, and set the learning rate for vertex offsets to $1\times 10^{-4}$. 
The Adam~\cite{kingma2014adam} optimizer is utilized for both stages. 
The coarse mesh is extracted at a resolution of $512^3$ with a density threshold of $10$, and its face number is decimated to $3 \times 10^{5}$ after extraction. 
We maintain a density grid to facilitate ray pruning, following the approach proposed in Instant-NGP~\cite{mueller2022instant}.
All experiments are conducted on a single NVIDIA V100 GPU.
Please refer to the supplementary materials for more details.

\noindent \textbf{Datasets.}
We experiment on three datasets to verify the effectiveness and generalization ability of our method:
1) NeRF-Synthetic~\cite{mildenhall2020nerf} dataset contains 8 synthetic scenes.
2) LLFF~\cite{mildenhall2019llff} dataset contains 8 realitic forward-facing scenes.
3) Mip-NeRF 360~\cite{barron2022mipnerf360} dataset contains 3 publicly available realistic unbounded outdoor scenes.
Our method generalize well to different types of datasets and reconstruct faithful mesh even for challenging unbounded scenes.

\subsection{Comparisons}

\subsubsection{Mesh Quality}

\noindent \textbf{Surface reconstruction.}
The lack of ground truth meshes for realistic scenes makes it challenging to measure surface reconstruction quality. 
As such, we primarily compare results on synthetic datasets, as done in NVdiffrec~\cite{munkberg2022extracting}. 
We provide qualitative assessments of the extracted meshes produced by different methods, as shown in Figure~\ref{fig:mesh_quality}. 
Specifically, we focus on thin structures such as dense foliage and rope net.
Our method successfully reconstructs these structures with high fidelity, while other methods fail to reconstruct the complex geometry accurately.
Additionally, our method produces meshes with more order and neatness, similar to human-made ground truths.

To quantify the surface reconstruction quality, we employ the bi-directional Chamfer Distance (CD) metric. 
However, as the ground truth meshes may not be surface meshes (\textit{e.g.}, the Lego mesh is actually made up of many small bricks), 
we cast rays from the test cameras and sample $2.5$M points from these ray-surface intersections per scene. 
In Table~\ref{tab:cd}, we present the averaged CD for all scenes, demonstrating that our method achieves the best results. 
We note that our approach performs particularly well on scenes with complex topology, such as ficus, ship, and lego. 
However, our method performs slightly worse on scenes with lots of non-lambertian surfaces, such as materials. 
This originates from the relatively limited capacity of our appearance network, as our model attempts to replicate these lighting effects by adjusting the surface but ends up with erroneous geometry.

\begin{table}[!t]
\begin{center}
\resizebox{1.0\linewidth}{!}{
\begin{tabular}{l|cc|cc|cc}
\hline
 & \multicolumn{2}{c|}{NeRF-synthetic} & \multicolumn{2}{c|}{LLFF} & \multicolumn{2}{c}{Mip-NeRF 360} \\
 & \#V & \#F & \#V & \#F & \#V & \#F \\
\hline
Ground Truth                                & 631  & 873   & -       & -       & -         & -       \\
\hline
NeuS~\cite{wang2021neus}                    & 1020 & 2039  & -       & -       & -         & -       \\
NVdiffrec~\cite{munkberg2022extracting}     & 75   & 80    & -       & -       & -         & -       \\
MobileNeRF~\cite{chen2022mobilenerf}        & 494  & 224   & 830 & 339 & 1436 & 609 \\
\hline
Ours (coarse mesh)                          & 151  & 300   & 231 & 455 & 446  & 886 \\
Ours (fine mesh)                            & 200  & 192   & 397 & 446 & 718  & 816 \\
\hline
\end{tabular}
}
\end{center}
\caption{
\textbf{Number of Vertices and Faces $\downarrow$ (Unit is $10^3$).} 
Our method uses relateively fewer vertices and faces on the NeRF-synthetic dataset with enhanced mesh quality.
}
\label{tab:v_t_count}
\end{table}
\begin{table}[!t]
\begin{center}
\resizebox{1.0\linewidth}{!}{
\begin{tabular}{l|cc|cc|cc}
\hline
 & \multicolumn{2}{c|}{NeRF-synthetic} & \multicolumn{2}{c|}{LLFF} & \multicolumn{2}{c}{Mip-NeRF 360} \\
 & Disk & Memory &  Disk & Memory &  Disk & Memory \\
\hline
SNeRG~\cite{hedman2021snerg}           & 86.75 & 2707.25 & 337.25 & 4312.13 & - & - \\
MobileNeRF~\cite{chen2022mobilenerf}   & 125.75 & 538.38 & 201.50 & 759.25 & 344.60 & 1080.00 \\
\hline
Ours                                   & 73.53 & 226.63 & 124.84 & 291.50 & 186.84 & 411.33 \\
\hline
\end{tabular}
}
\end{center}
\caption{
\textbf{Disk Storage and GPU Memory Usage $\downarrow$ ({MB}).} 
We measure the size of exported models and GPU memory usage in rendering.
}
\label{tab:storage}
\end{table}

\noindent \textbf{Mesh size.}
We also evaluate the practical applicability by comparing the number of vertices and faces in the exported meshes, as shown in Table~\ref{tab:v_t_count}. 
Furthermore, we measure the disk storage and GPU memory usage required for rendering the exported meshes, as presented in Table~\ref{tab:storage}. 
For fair comparison, the mesh file format is uncompressed OBJ \& MTL, the texture is PNG, and other metadata is stored in JSON format. 
Compared to the ground truth and MobileNeRF~\cite{chen2022mobilenerf} on the NeRF-synthetic dataset, our exported meshes contain fewer vertices and faces. 
This is because the iterative mesh refinement process can increase the number of vertices to enhance surface details, while simultaneously reducing the number of faces to control mesh size. 


\begin{figure}[t]
    \centering
    \includegraphics[width=\linewidth]{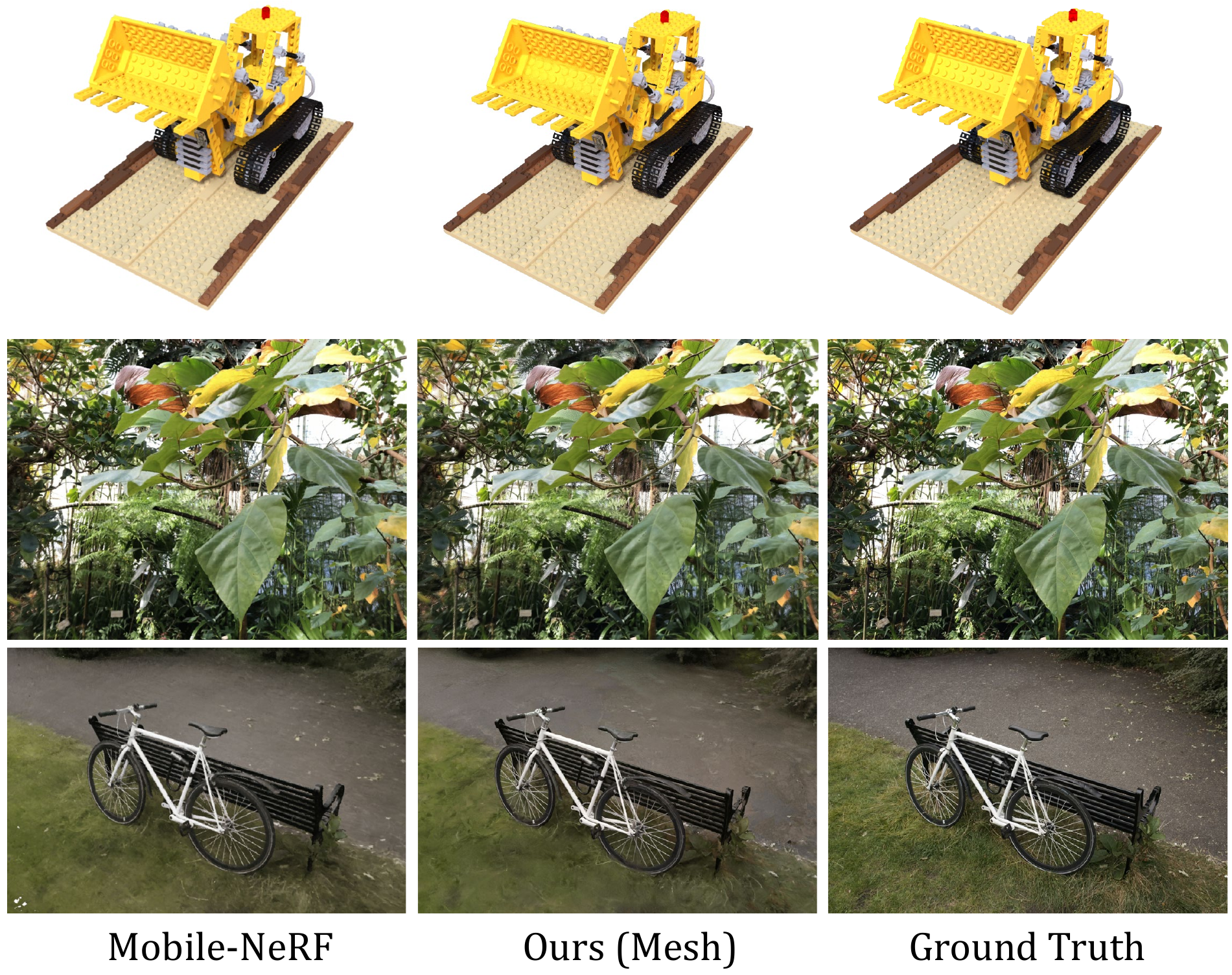}
    \caption{
    \textbf{Visualization of rendering quality}. Our exported meshes achieve comparable rendering quality on different datasets.
    }
    \vspace{-0.2cm}
    \label{fig:rendering_quality}
\end{figure}

\begin{figure}[t]
    \centering
    \includegraphics[width=\linewidth]{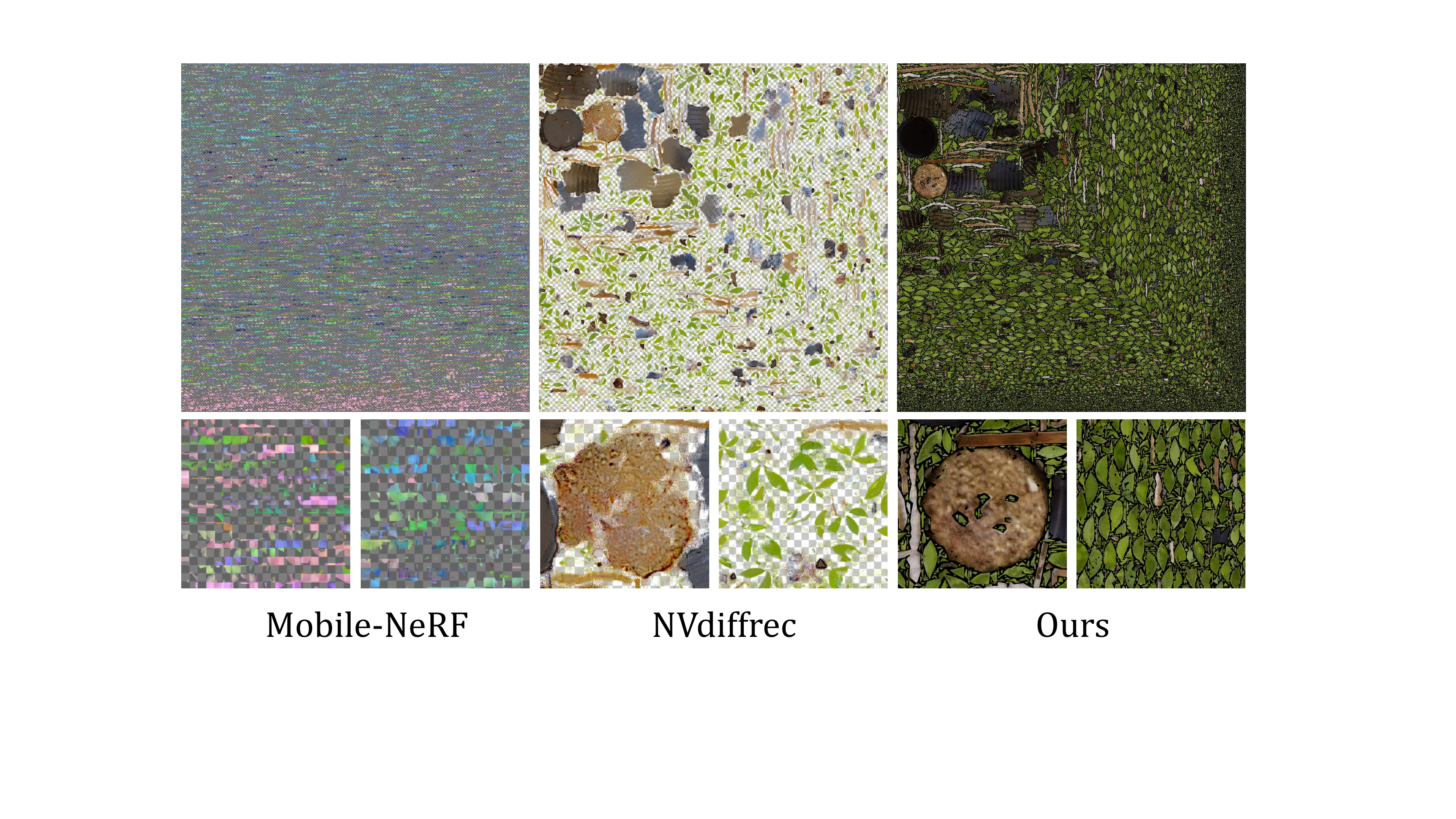}
    \caption{
    \textbf{Visualization of texture images}. We show that our textures are more compact and intuitive due to the enhanced surface quality.
    }
    \vspace{-0.2cm}
    \label{fig:texture_quality}
\end{figure}

\begin{table*}[!t]
\begin{center}
\resizebox{1.0\linewidth}{!}{
\begin{tabular}{l|c|ccc|ccc|ccc}
\hline
 & \multirow{ 2}{*}{Category} & \multicolumn{3}{c|}{NeRF-synthetic} & \multicolumn{3}{c|}{LLFF} & \multicolumn{3}{c}{Mip-NeRF 360} \\
 &  & PSNR$\uparrow$ & SSIM$\uparrow$ & LPIPS$\downarrow$ & PSNR$\uparrow$ & SSIM$\uparrow$ & LPIPS$\downarrow$ & PSNR$\uparrow$ & SSIM$\uparrow$ & LPIPS$\downarrow$ \\
\hline
NeRF~\cite{mildenhall2020nerf}         & \multirow{ 2}{*}{Volume} & 31.00 & 0.947 & 0.081 & 26.50 & 0.811 & 0.250 & - & - & - \\
Ours (volume)                          &                          & 30.88 & 0.951 & 0.079 & 26.42 & 0.824 & 0.218 & 22.33 & 0.538 & 0.481 \\
\hline
NVdiffrec~\cite{munkberg2022extracting}& Surface                  & 29.05 & 0.939 & 0.081 & - & - & - & - & - & - \\
Ours (mesh)                            & mesh                     & 29.76 & 0.940 & 0.072 & 24.75 & 0.780 & 0.267 & 22.36 & 0.493 & 0.478 \\
\hline
MobileNeRF~\cite{chen2022mobilenerf}   & Non-surface              & 30.90 & 0.947 & 0.062 & 25.91 & 0.825 & 0.183 & 23.06 & 0.527 & 0.434 \\
Ours (mesh w/o $\mathcal L_\text{smooth}$)    & mesh              & 31.04 & 0.948 & 0.066 & 24.90 & 0.778 & 0.271 & 22.74 & 0.523 & 0.457 \\
\hline
\end{tabular}
}
\end{center}
\caption{\textbf{Rendering Quality Comparison.} 
We report PSNR, SSIM, and LPIPS on different datasets, and compare against methods from different categories. 
}
\label{tab:render_quality}
\end{table*}

\subsubsection{Rendering Quality}

We present the results of our rendering quality comparison in Table~\ref{tab:render_quality}. 
We observe a decrease in rendering quality from NeRF (volume) to mesh. 
Specifically, we find that the smoothness regularization term $\mathcal L_\text{smooth}$ plays a crucial role in maintaining a balance between surface smoothness and rendering quality. 
Disabling this regularization term leads to better rendering quality at the expense of surface quality (detailed in Section~\ref{sec:ablation}).
We demonstrate that our mesh-based approach yields superior rendering quality compared to NVdiffrec~\cite{munkberg2022extracting}, which is state-of-the-art in the surface mesh category. 
Furthermore, our approach generalizes well to forward-facing and unbounded scenes, while NVdiffrec~\cite{munkberg2022extracting} is only capable of reconstructing single objects.
MobileNeRF~\cite{chen2022mobilenerf} exports grid-like meshes that lack smoothness and may not align well with object surfaces. 
These meshes rely on texture transparency to carve out the surface. 
Although our smooth meshes exhibit worse rendering quality, our meshes without the smoothness regularization term achieve comparable performance.
Figure~\ref{fig:rendering_quality} presents a visualization of our meshes' rendering quality and compares them with related methods.
In Figure~\ref{fig:texture_quality}, we also present the texture images exported by different methods. 
We demonstrate that our high-quality surface meshes result in texture images that are more compact and intuitive than those generated by other methods.

\subsection{Efficiency}

Our framework demonstrates high efficiency in both training and inference stages. 
A single NVIDIA V100 GPU with 16GB memory takes roughly 1 hour for the training of the two stages and mesh exportation per scene. 
In contrast, other competing methods often require several hours~\cite{munkberg2022extracting} or even days~\cite{chen2022mobilenerf}, with higher hardware demands to complete similar tasks. 
Furthermore, the exported meshes are lightweight, allowing for real-time rendering on OpenGL-enabled devices, including mobile devices.

\begin{figure}[t]
    \centering
    \includegraphics[width=\linewidth]{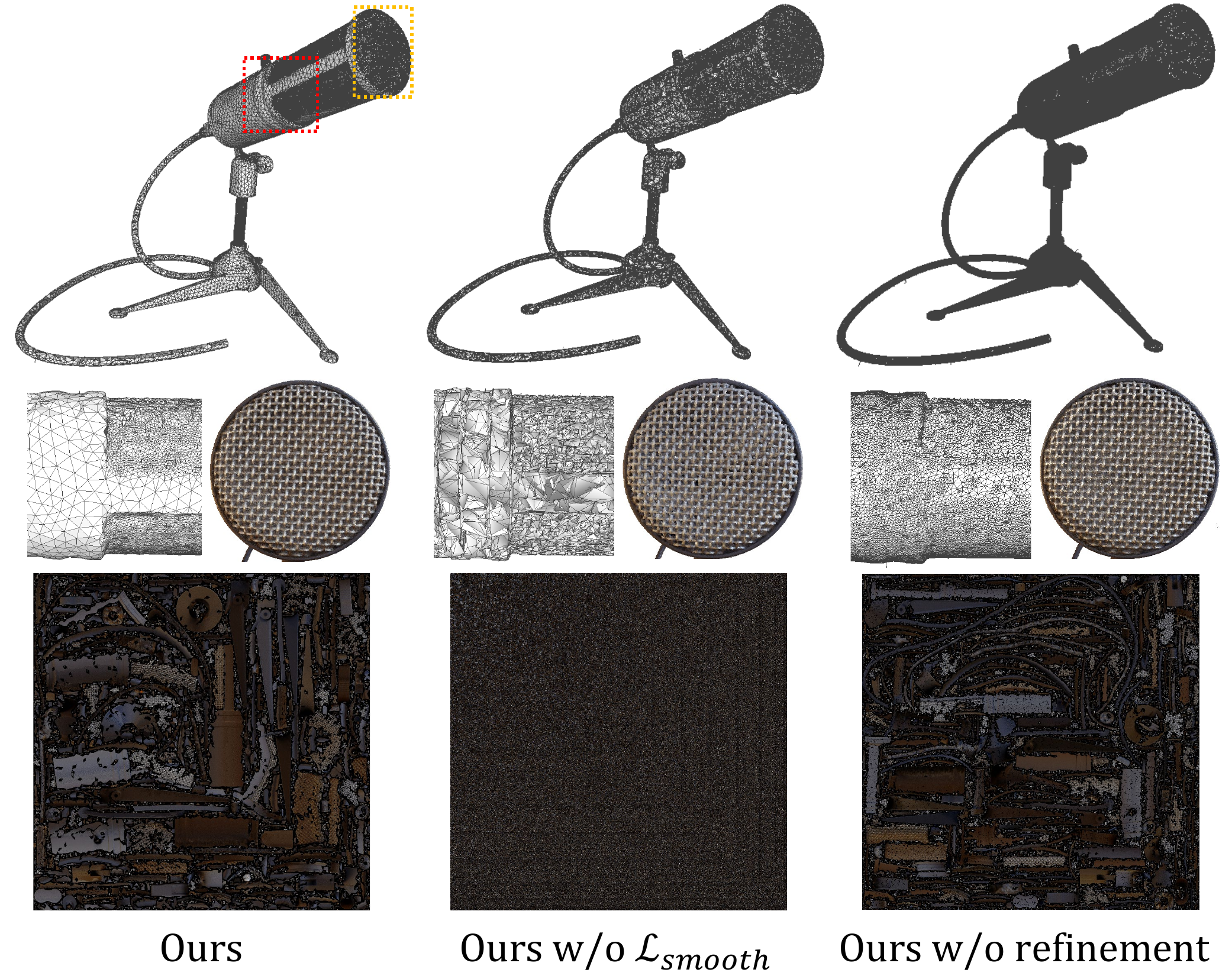}
    \caption{
    \textbf{Qualitative Ablation.} We visualize the mesh structure and texture images for the Mic scene.
    }
    \label{fig:ablation}
\end{figure}

\subsection{Ablation Studies}
\label{sec:ablation}

In Figure~\ref{fig:ablation} and Table~\ref{tab:ablation}, we conduct an ablation study focusing on the geometry optimization stage. 
Specifically, we compare the full model against variants that exclude either the smoothness regularization $\mathcal L_\text{smooth}$ or the iterative mesh refinement process. 
The results indicate that:
1) When the smoothness regularization is removed, despite the better rendering quality, the resulting mesh exhibits irregularities and self-intersections. 
The mesh size increases due to the failure of the iterative mesh refinement algorithm to work well with such irregular surfaces. 
Moreover, these irregular faces lead to poor UV quality and messy texture images.
2) When the iterative mesh refinement is removed, the face density becomes nearly uniform, resulting in a larger mesh size and slightly inferior rendering quality.
This illustrates the advantages of modifying face density according to the errors in the re-projected rendering.

\begin{table}[!t]
\begin{center}
\resizebox{1.0\linewidth}{!}{
\begin{tabular}{l|cccc}
\hline
                                           & \#V  & \#F & Size (MB) & PSNR \\
\hline

Ours                                       & 58,649  & 116,698 & 54.8  & 31.30 \\
Ours w/o $\mathcal L_\text{smooth}$        & 202,656 & 396,385 & 133.0 & 32.57 \\
Ours w/o refinement                        & 150,276 & 300,000 & 74.8  & 31.06 \\
\hline
\end{tabular}
}
\end{center}
\caption{
\textbf{Quantitative Ablation.} We report the mesh statistics and PSNR on the Mic scene.
}
\label{tab:ablation}
\end{table}

\section{Limitations and Conclusion}
Although our method has shown promising results, it still has several limitations. 
Due to the difficulty of estimating unknown lighting conditions from images without compromising reconstruction quality~\cite{zhang2021nerfactor}, we have chosen to bake illumination into textures, which consequently restricts our ability to perform relighting. 
Our relatively small appearance network also struggles to learn complex view-dependent effects, confounding iterative surface refinement and resulting in inferior surface quality within these regions.
In the future, we hope to address these limitations by leveraging better appearance modeling techniques. 
Lastly, similar to other mesh-based methods~\cite{munkberg2022extracting,chen2022mobilenerf}, we perform a single-pass rasterization and are unable to handle semi-transparency.

In summary, we present an efficient framework that can reconstruct textured surface meshes from multi-view RGB images. 
Our approach utilizes NeRF for coarse geometry and appearance initialization, subsequently extracts and enhances a polygonal mesh, and ultimately bakes the appearance into texture images for real-time rendering. 
The reconstructed meshes demonstrate an enhanced surface quality, particularly for thin structures, and are convenient to manipulation and editing fow downstream applications.

\clearpage
\noindent \textbf{Acknowledgements.} 
This work is supported by the Sichuan Science and Technology Program (2023YFSY0008), National Natural Science Foundation of China (61632003, 61375022, 61403005), Grant SCITLAB-20017 of Intelligent Terminal Key Laboratory of SiChuan Province, Beijing Advanced Innovation Center for Intelligent Robots and Systems (2018IRS11), and PEK-SenseTime Joint Laboratory of Machine Vision.

\begin{appendix}
\appendix

\section{Additional Implementation Details}

\noindent \textbf{Network Architecture.}
We use the multi-resolution hashgrid encoder~\cite{mueller2022instant} and shallow MLPs to cosntruct the first stage's NeRF network.
The density grid $E^\text{geo}$ use 16 resolution levels with each level containing 1-channel features, and a 2-layer MLP with 32 hidden channels is used to convert the features into density.
The color grid $E^\text{app}$ use 16 resolution levels with each level containing 2-channel features.
A 3-layer MLP with 64 hidden channels convert the color features into 3-channel diffuse color and 3-channel specular features.
The specular features along with view directions are fed into a 2-layer MLP with 32 hidden channels to produce the view-dependent 3-channel specular color.

\noindent \textbf{Visibility culling \& Mesh cleaning.}
In the first stage of our approach, we adopt the Marching Cubes algorithm to extract a coarse mesh from NeRF's density field. 
To reduce the size of the resultant mesh, we incorporate a visibility culling mechanism to eliminate vertices and faces that are invisible from all training cameras. 
More specifically, we cast rays from each training camera and calculate their intersection with the surface. 
In doing so, we trace the corresponding face and label it as visible. 
However, in situations where the training cameras are sparsely located, this approach may result in excessive culling. 
To address this issue, we apply dilation to the visible faces using a predetermined kernel size. 
For the NeRF-synthetic dataset, we utilize a kernel size of 5, while for the LLFF and Mip-NeRF 360 dataset, we increase it to a larger value of 50, given that training cameras may be sparse for the far background. 
The mesh can be further post-processed to remove floaters based on the diameter and number of faces for each connected component.
We also clean the mesh by merging close vertices, removing duplicated faces, and repairing non-manifold vertices and faces~\cite{meshlab}.
In essence, these methods help to remove unnecessary vertices and faces to maintain a reasonably small mesh size.

\noindent \textbf{Baking.}
After completing the two-stage training, we convert the appearance network into texture images for real-time rendering. 
Initially, the resolution of the texture image is set to 4096 for the center mesh in $[-1, 1]^3$. 
Subsequently, for meshes of outdoor regions, the texture resolution is decreased by a power of 2, with a minimum resolution of 1024. 
To eliminate seam-like texture artifacts caused by UV unwrapping~\cite{munkberg2022extracting}, we repair the border of each connected component by out-painting 1 pixel on the texture image. 
The floating-point diffuse color and specular features in $[0, 1]$ range are quantized into 8-bit precision PNG images. 
Following MobileNeRF~\cite{chen2022mobilenerf}, we found that the rendering quality is not significantly affected through baking.

\noindent \textbf{Hyper-parameters.}
Since different types of dataset (\textit{e.g.}, from objects without background to unbounded scenes) can require very different hyper-parameters to maximize performance~\cite{mildenhall2020nerf,sun2021direct}, we explore different set of hyper-parameters, especially for loss weights.
By default, we set the weight of $\mathcal L_\text{TV}$ to $1\times 10^{-8}$, the weight of $\mathcal L_\text{smooth}$ to $1\times 10^{-3}$, and the weight of $\mathcal L_\text{offset}$ to $0.1$.
The other loss weights are default to $0$ unless specified.
For the NeRF-synthetic dataset and the LLFF dataset, we use all the default weights.
For the Mip-NeRF 360 dataset, we set the weight of $\mathcal L_\text{entropy}$ to $1\times 10^{-3}$.
The training steps for stage 1 is also set differently.
We train $30,000$ steps for the NeRF-synthetic dataset, but we found $10,000$ steps are enough for the LLFF and Mip-NeRF 360 datasets to converge.
For the iterative mesh refining algorithm, we apply the subdivision and decimation at $\{0.1, 0.2, 0.3, 0.4, 0.5, 0.7\}$ ratio of total training steps.
The minimum edge length for subdivision is set to 1\% of the diagonal of the bounding box of the mesh (which eqauls to $0.02 \sqrt 3$ in our case).
We decimate 10\% of the faces with an error above $e_\text{decimate}$, and remesh them with an average edge length of 2\% of the diagonal of the bounding box of the mesh ($0.04 \sqrt 3$).

\begin{figure}[ht]
    \centering
    \includegraphics[width=\linewidth]{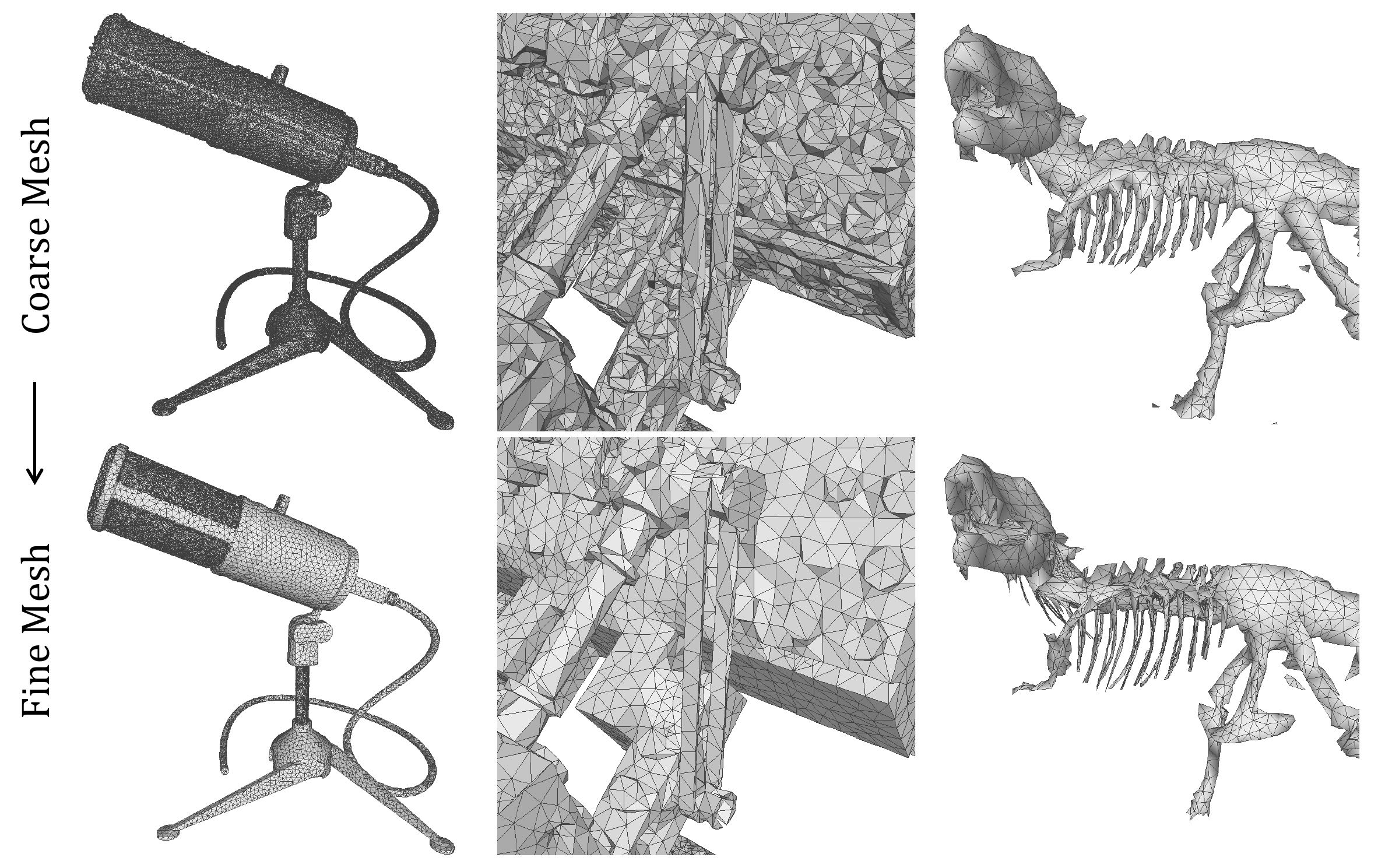}
    \caption{
    \textbf{Adaptive Face Density}.
    Our iterative mesh refining allows adaptive face density learned from data. It enhances the surface quality and reduces face counts.
    }
    \label{fig:refine_face}
\end{figure}

\begin{figure}[ht]
    \centering
    \includegraphics[width=\linewidth]{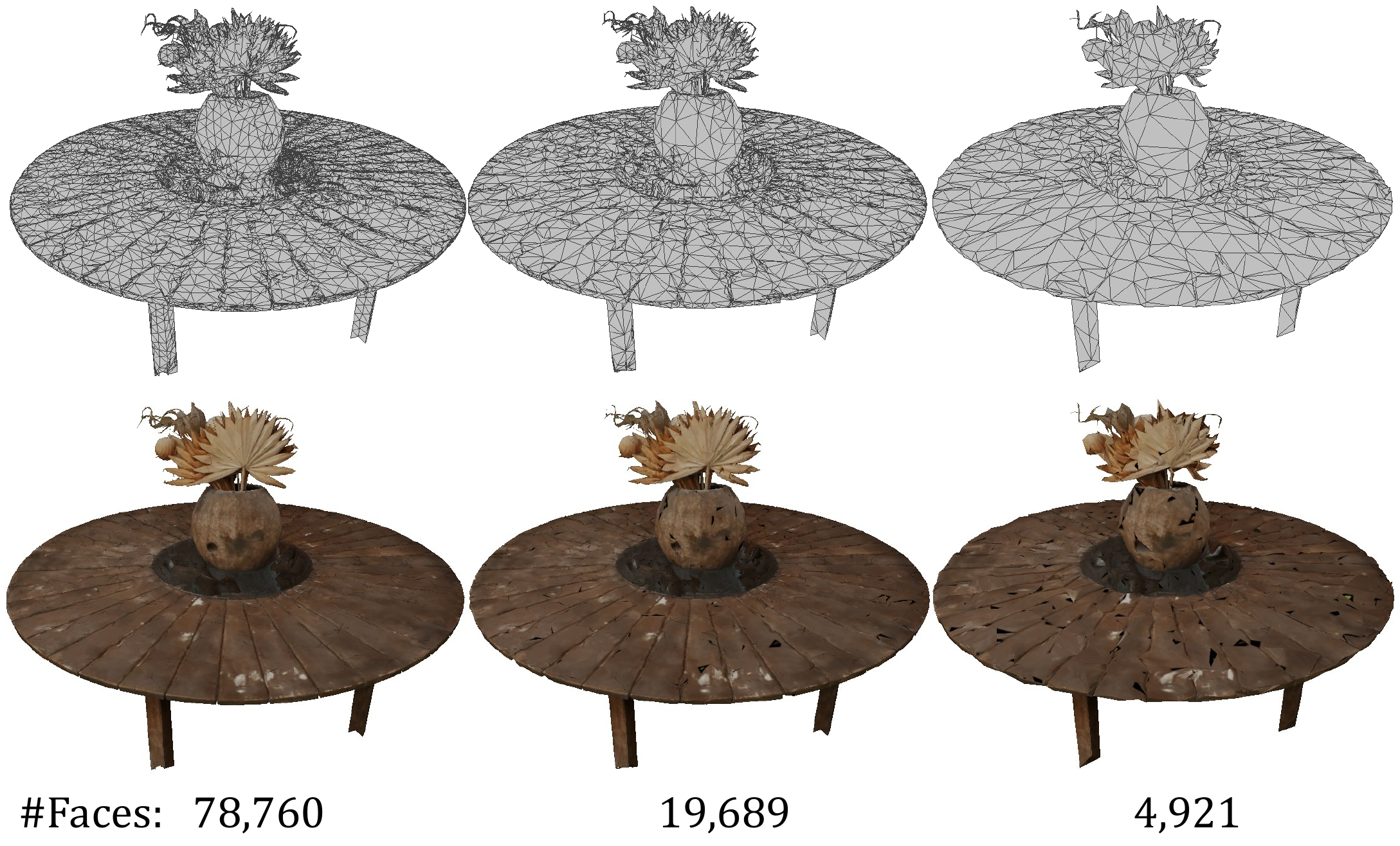}
    \caption{
    \textbf{Levels of details (LOD) simulation}. 
    We decimate the reconstructed mesh to create different LODs.
    }
    \label{fig:lod}
\end{figure}

\begin{figure}[ht]
    \centering
    \includegraphics[width=\linewidth]{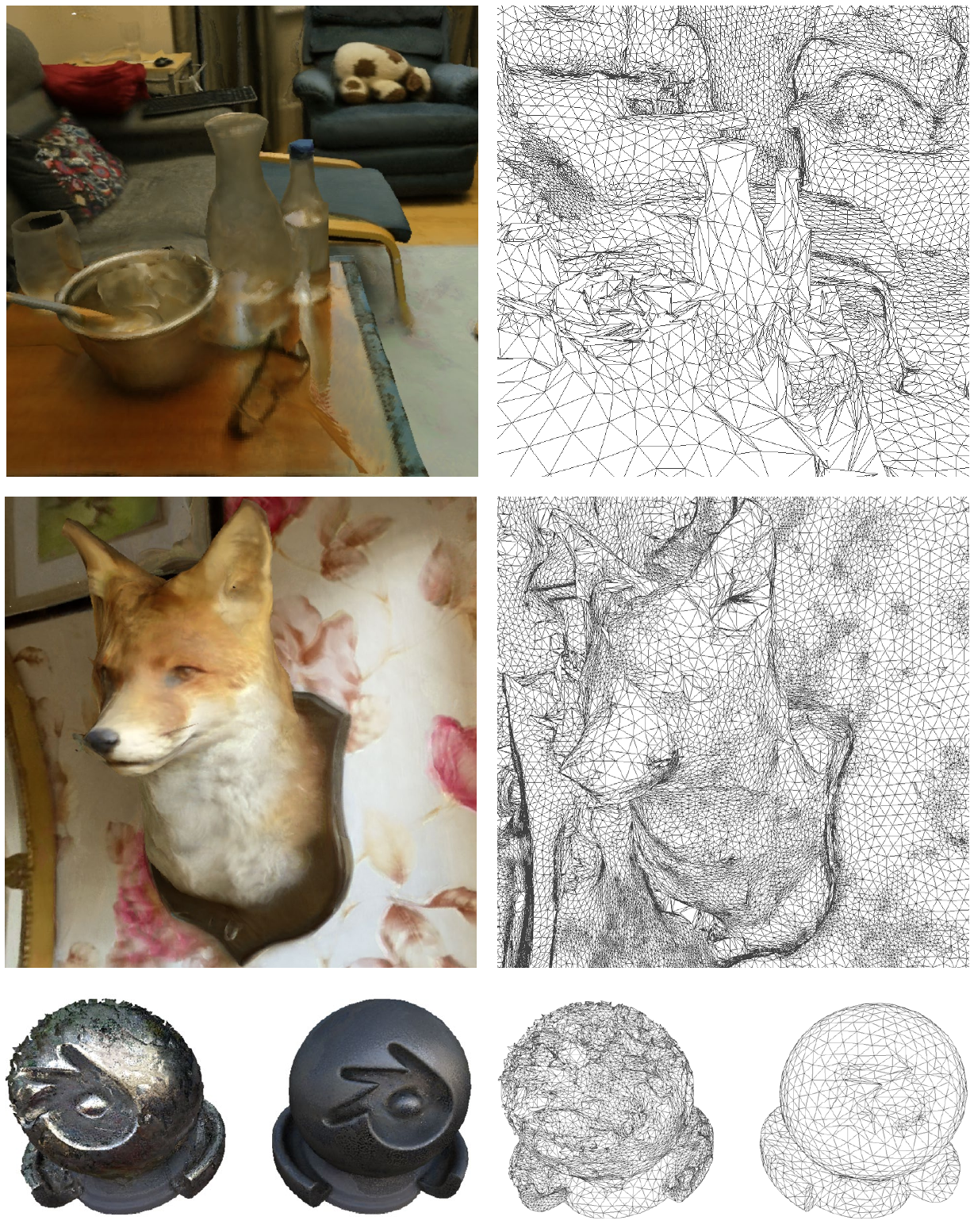}
    \caption{
    \textbf{Limitations}. 
    We visualize some examples about the limitation of our method.
    }
    \label{fig:limitation}
\end{figure}

\begin{figure}[ht]
    \centering
    \includegraphics[width=\linewidth]{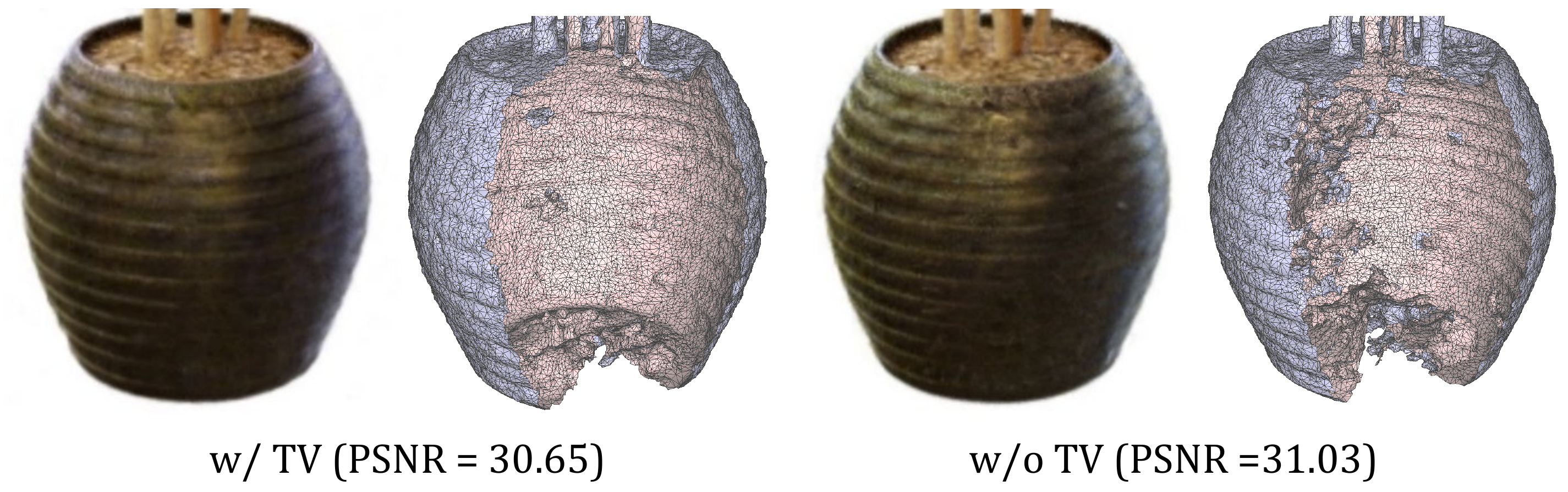}
    \caption{Ablation on TV loss. We remove some surface to show internal geometry.}
    \label{fig:abl_tv}
\end{figure}

\begin{figure*}[ht]
    \centering
    \includegraphics[width=\linewidth]{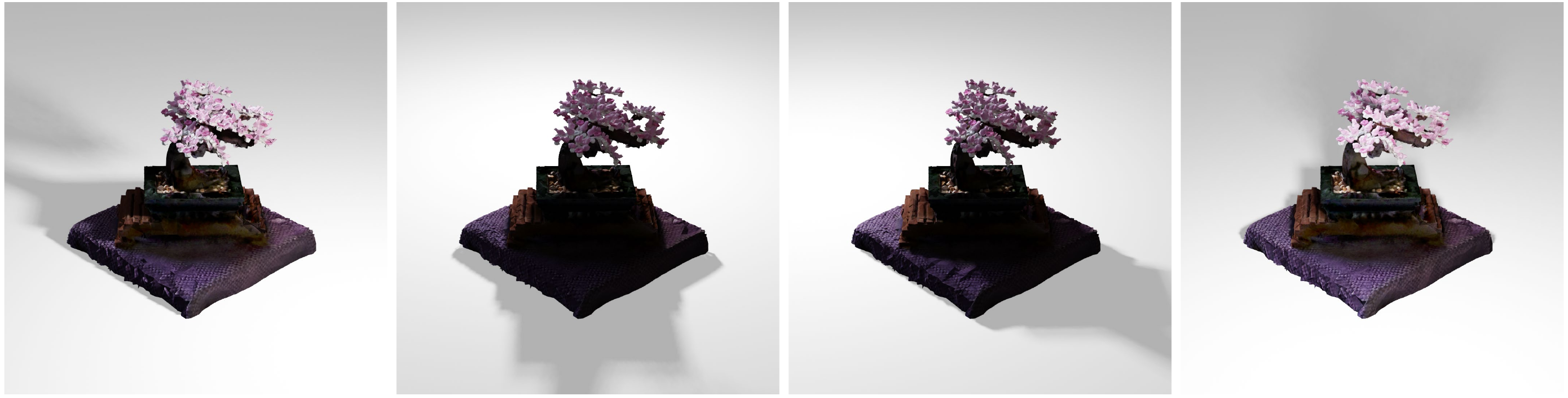}
    \caption{
    \textbf{Relighting}. 
    We relight the mesh with only the diffuse texture with a point light source.
    }
    \label{fig:relight}
\end{figure*}

\begin{figure*}[ht]
    \centering
    \includegraphics[width=\linewidth]{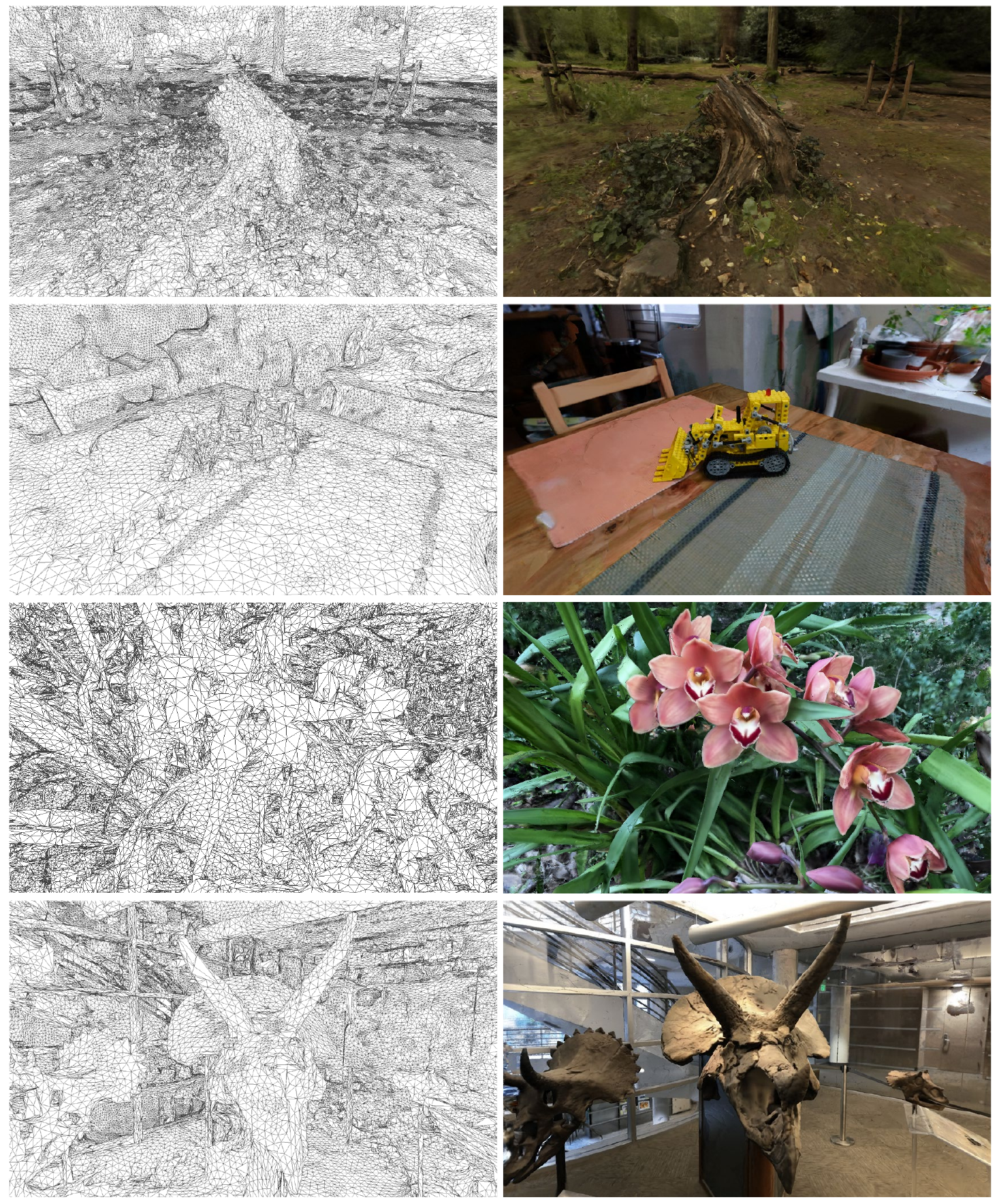}
    \caption{
    \textbf{More Visualizations}. 
    We visualize the mesh and diffuse color of more scenes from the Mip-NeRF 360 and LLFF datasets.
    }
    \label{fig:more_360}
\end{figure*}

\section{Additional Experimental Results}

\subsection{Additional Qualitative Results}

\noindent \textbf{Relighting.}
Although the lighting is baked into textures in our methods, we aim to showcase that our mesh is proficient enough to execute relighting for a scene captured in predominantly ambient lighting conditions. 
Figure~\ref{fig:relight} exhibits a reconstructed mesh which has been relit with a rotating point light source. Only the diffuse texture is utilized.

\noindent \textbf{Levels of detail (LODs) simulation.}
Our textured surface mesh is demonstrated to be suitable for supporting LODs in Figure~\ref{fig:lod}. 
This can solely be accomplished with an accurate surface mesh, as decimation is primarily designed for minimizing geometric error.

\noindent \textbf{Additional Visualizations.}
We also provide more visualization on scenes with background in Figure~\ref{fig:more_360}.
In Figure~\ref{fig:refine_face}, we show more visualizations on the iterative mesh refining.
In Figure~\ref{fig:abl_tv}, we visualize the effect of the TV loss on mesh quality. 

\noindent \textbf{Limitations.}
Our method's limitations are illustrated in Figure~\ref{fig:limitation}. 
As we solely perform single-layer rasterization, our approach is incapable of handling semi-transparent objects such as glass bottles, and tends to learn an opaque texture. 
Animal fur, which usually requires volumetric representation for better simulation, is difficult to emulate due to the smoothness regularization of the mesh surface. 
Lastly, since the appearance network is relatively small, it cannot model intricate view-dependent effects. 
Therefore, our model tends to manipulate vertices to simulate the effects, which results in a lack of smoothness and inaccurate geometry. 
We are hopeful for improved decomposition of surface and materials to overcome this issue.

\subsection{Additional Quantitative Results}

\begin{table}[t]
\begin{center}
\begin{tabular}{l|ccc}
\hline
                                           & PSNR$\uparrow$  & SSIM$\uparrow$ & LPIPS$\downarrow$ \\
\hline

Ours                                       & 22.36 & 0.493 & 0.478  \\
Ours w/o $\mathcal L_\text{entropy}$       & 22.32 & 0.492 & 0.481  \\
Ours w/o $\mathcal L_\text{TV}$            & 22.22 & 0.486 & 0.483  \\
\hline
\end{tabular}
\end{center}
\caption{We ablate the regularizations on the Mip-NeRF 360 dataset.}
\label{tab:more_ablation}
\end{table}

The per-scene rendering quality evaluation results are listed in Table~\ref{table:render_syn}, Table~\ref{table:render_llff}, and Table~\ref{table:render_360}.
In Table~\ref{tab:more_ablation}, we perform more ablation on the regularization losses.

\begin{table*}[!t]
\begin{center}
\resizebox{1.0\linewidth}{!}{
\begin{tabular}{l|c|cccccccc|c}
\hline
& Metric                                                                  & Chair & Drums & Ficus & Hotdog & Lego & Materials & Mic & Ship & Mean \\
\hline
Ours (volume)                           & \multirow{3}{*}{PSNR$\uparrow$} & 33.87 & 25.20 & 30.24 & 35.09 & 33.66 & 27.70 & 32.65 & 28.59 & 30.88 \\
Ours (mesh)                             &                                 & 31.93 & 24.80 & 29.81 & 34.11 & 32.07 & 25.45 & 31.25 & 28.69 & 29.76 \\
Ours (mesh w/o $\mathcal L_\text{smooth}$)                   &            & 34.25 & 25.04 & 30.08 & 35.70 & 34.90 & 26.26 & 32.63 & 29.47 & 31.04 \\
\hline
Ours (volume)                           & \multirow{3}{*}{SSIM$\uparrow$} & 0.977 & 0.929 & 0.970 & 0.974 & 0.972 & 0.930 & 0.981 & 0.872 & 0.951 \\
Ours (mesh)                             &                                 & 0.964 & 0.927 & 0.967 & 0.970 & 0.957 & 0.896 & 0.974 & 0.865 & 0.940 \\
Ours (mesh w/o $\mathcal L_\text{smooth}$)                   &            & 0.978 & 0.926 & 0.967 & 0.974 & 0.977 & 0.906 & 0.979 & 0.875 & 0.948 \\
\hline
Ours (volume)                        & \multirow{3}{*}{LPIPS$\downarrow$} & 0.049 & 0.108 & 0.064 & 0.052 & 0.055 & 0.091 & 0.047 & 0.163 & 0.079 \\
Ours (mesh)                             &                                 & 0.046 & 0.084 & 0.045 & 0.060 & 0.047 & 0.107 & 0.042 & 0.145 & 0.072 \\
Ours (mesh w/o $\mathcal L_\text{smooth}$)                   &            & 0.031 & 0.084 & 0.046 & 0.058 & 0.025 & 0.111 & 0.038 & 0.138 & 0.066 \\
\hline
\end{tabular}
}
\end{center}
\vspace{-0.3cm}
\caption{
Rendering quality on the NeRF-synthetic dataset.
}
\label{table:render_syn}
\end{table*}

\begin{table*}[!t]
\begin{center}
\resizebox{1.0\linewidth}{!}{
\begin{tabular}{l|c|cccccccc|c}
\hline
 & Metric                                                              & Room & Fern & Leaves & Fortress & Orchids & Flower & Trex & Horns & Mean \\
\hline
Ours (volume)                        & \multirow{3}{*}{PSNR$\uparrow$} & 31.12 & 25.47 & 20.58 & 30.45 & 20.54 & 27.19 & 28.15 & 27.83 & 26.42 \\
Ours (mesh)                          &                                 & 29.24 & 23.94 & 19.22 & 28.02 & 19.08 & 26.48 & 25.80 & 26.25 & 24.75 \\
Ours (mesh w/o $\mathcal L_\text{smooth}$)                &            & 30.03 & 23.21 & 18.71 & 28.96 & 19.34 & 26.10 & 26.41 & 26.40 & 24.90 \\
\hline
Ours (volume)                        & \multirow{3}{*}{SSIM$\uparrow$} & 0.939 & 0.802 & 0.700 & 0.887 & 0.668 & 0.823 & 0.910 & 0.866 & 0.824 \\
Ours (mesh)                          &                                 & 0.914 & 0.751 & 0.644 & 0.765 & 0.602 & 0.879 & 0.868 & 0.819 & 0.780 \\
Ours (mesh w/o $\mathcal L_\text{smooth}$)                &            & 0.923 & 0.709 & 0.621 & 0.859 & 0.607 & 0.797 & 0.879 & 0.831 & 0.778 \\
\hline
Ours (volume)                    & \multirow{3}{*}{LPIPS$\downarrow$}  & 0.201 & 0.248 & 0.253 & 0.167 & 0.270 & 0.204 & 0.180 & 0.223 & 0.218 \\
Ours (mesh)                          &                                 & 0.246 & 0.303 & 0.321 & 0.270 & 0.314 & 0.204 & 0.215 & 0.260 & 0.267 \\
Ours (mesh w/o $\mathcal L_\text{smooth}$)                &            & 0.254 & 0.342 & 0.358 & 0.203 & 0.312 & 0.224 & 0.214 & 0.259 & 0.271 \\
\hline
\end{tabular}
}
\end{center}
\vspace{-0.3cm}
\caption{
Rendering quality on the LLFF dataset.
}
\label{table:render_llff}
\end{table*}

\begin{table*}[!t]
\begin{center}

\begin{tabular}{l|c|ccc|c}
\hline
                          & Metric                                      & Bicycle    & Garden   & Stump & Mean \\
\hline
Ours (volume)                        & \multirow{3}{*}{PSNR$\uparrow$}  & 20.88      & 23.41 & 22.70      & 22.33 \\
Ours (mesh)                          &                                  & 22.16      & 22.39 & 22.53      & 22.36 \\
Ours (mesh w/o $\mathcal L_\text{smooth}$)                &             & 22.28      & 22.86 & 23.08      & 22.74 \\
\hline
Ours (volume)                        & \multirow{3}{*}{SSIM$\uparrow$}  & 0.469      & 0.567 & 0.578      & 0.538 \\
Ours (mesh)                          &                                  & 0.470      & 0.500 & 0.508      & 0.493 \\
Ours (mesh w/o $\mathcal L_\text{smooth}$)                &             & 0.479      & 0.551 & 0.540      & 0.523 \\
\hline
Ours (volume)                    & \multirow{3}{*}{LPIPS$\downarrow$}   & 0.545      & 0.419 & 0.478      & 0.481 \\
Ours (mesh)                          &                                  & 0.510      & 0.434 & 0.490      & 0.478 \\
Ours (mesh w/o $\mathcal L_\text{smooth}$)                &             & 0.509      & 0.402 & 0.459      & 0.457 \\
\hline
\end{tabular}

\end{center}
\vspace{-0.3cm}
\caption{
Rendering quality on the Mip-NeRF 360 dataset.
}
\label{table:render_360}
\end{table*}

\end{appendix}

{\small
\bibliographystyle{ieee_fullname}
\bibliography{06_ref}
}

\end{document}